\documentclass[10pt,conference,compsocconf]{IEEEtran}
\usepackage{times}

\usepackage{caption}
\usepackage[noadjust]{cite}
\usepackage{amsmath,amssymb,amsfonts}
\usepackage{algorithmic}
\usepackage{graphicx}
\usepackage{textcomp}
\usepackage{xcolor}

\ifCLASSINFOpdf
\else
\fi
\begin{document}
	\pagenumbering{gobble}
	\title{\textbf{\Large Efficient Intrusion Detection Using Evidence Theory}\\[0.2ex]}
	

	\author{\IEEEauthorblockN{~\\[-0.4ex]\large Islam Debicha\\[0.3ex]\normalsize}
	\IEEEauthorblockA{Royal Military Academy \\
		and Université Libre de Bruxelles\\
		Brussels, Belgium\\
	   Email: {\tt debichasislam@gmail.com}  }
	\and
	\IEEEauthorblockN{~\\[-0.4ex]\large Thibault Debatty\\[0.3ex]\normalsize}
	\IEEEauthorblockA{Royal Military Academy\\
		Brussels, Belgium\\
	\qquad \qquad	Email: {\tt thibault.debatty@rma.ac.be} \qquad \qquad}
	\and
	\IEEEauthorblockN{~\\[-0.4ex]\large Wim Mees\\[0.3ex]\normalsize}
	\IEEEauthorblockA{Royal Military Academy\\
		Brussels, Belgium\\
	\qquad \qquad \qquad	Email: {\tt wim.mees@rma.ac.be}  \qquad \qquad \qquad}
	\and
	\IEEEauthorblockN{~\\[-0.4ex]\large Jean-Michel Dricot\\[0.3ex]\normalsize}
	\IEEEauthorblockA{Université Libre de Bruxelles\\
		Brussels, Belgium\\
		Email: {\tt jdricot@ulb.ac.be}}
	
}
	
	
	%Telephone: (800) 555--1212, Fax: (888) 555--1212}


	\maketitle

	\begin{abstract}
		%\boldmath
		Intrusion Detection Systems (IDS) are now an essential element when it comes to securing computers and networks. Despite the huge research efforts done in the field, handling sources' reliability remains an open issue. To address this problem, this paper proposes a novel contextual discounting method based on sources' reliability and their distinguishing ability between normal and abnormal behavior. Dempster-Shafer theory, a general framework for reasoning under uncertainty, is used to construct an evidential classifier. The NSL-KDD dataset, a significantly revised and improved version of the existing KDDCUP'99 dataset, provides the basis for assessing the performance of our new detection approach. While giving comparable results on the KDDTest+ dataset, our approach outperformed some other state-of-the-art methods on the KDDTest-21 dataset which is more challenging.
	\end{abstract}
	
	

	% \fi
	\IEEEpeerreviewmaketitle

	\section{Introduction}
As computer network usage grows rapidly along with the significant increase in the number of applications running on it, the importance of network security is increasing. As dedicated tools designed to identify anomalies and attacks on the network, Intrusion Detection Systems (IDS) are becoming more valuable. Detection techniques based on anomalies and misuse have long been the principal subject of research in the field of intrusion detection \cite{andress2014basics}.

Misuse-based IDSs operate quite similarly to most antivirus systems. Maintaining a signature database that could identify specific types of attacks and checking all incoming traffic against these signatures. Overall, this approach performs well, although it does not work properly when dealing with new attacks, or those that were specifically crafted to mismatch existing signatures.

On the other hand, anomaly-based IDSs operate generally on a baseline of normal activities and network traffic. This allows them to assess the current state of network traffic against this baseline so that abnormal patterns can be identified. While such an approach could be quite effective in detecting new attacks or those that have been intentionally crafted to evade IDSs, it can also result in a higher number of false positives compared to misuse-based IDSs.

%Data fusion enhances the process of decision making when heterogeneous data sources are in place, primarily by leveraging duplication and complementarity across sources. 

Dempster-Shafer Theory (DST), also known as evidence theory \cite{shafer1976mathematical} is one of the most versatile mathematical frameworks, extending Bayesian theory by (i) providing each source with the ability to integrate information at various scales of detail, thus addressing uncertainty; and (ii) offering a robust decision-making tool to make a consensus-based decision. This theory was later widely applied in several domains \cite{jones2002framework}\cite{boudaren2016ijar}\cite{guo2006evaluating}. Regardless of this popularity, mass function generation and source reliability estimation remain an ongoing challenge. 

Probabilistic frameworks for mass generation take advantage of the extensive research literature of the traditional probabilistic classifiers.
These approaches usually represent the information associated with each attribute through Probability Density Functions (PDF), typically Gaussian \cite{salzenstein2001unsupervised}\cite{bendjebbour2001}. Such densities are then transformed into beliefs that can subsequently be merged to form a joint decision. One can attribute masses to the compound hypotheses by subtracting the mass values related to the simple hypotheses involved \cite{salzenstein2001unsupervised} or by mixing the distributions associated with these hypotheses \cite{bendjebbour2001}. It should be noted that for most applications, Gaussian densities have been widely assumed due to their simplicity. Nevertheless, in the case where this assumption fails, the decision-making performance may be influenced considerably. More sophisticated approaches can be used to surmount this limitation by transforming data attributes into an equivalent normal space \cite{xu2013new}.

This paper offers a more effective way to overcome this disadvantage by constructing PDFs that are better suited to the original data histograms instead of projecting them into a new Gaussian-like space. On a more explicit level, a kernel smoothing estimation \cite{wand1994kernel} is used on the training data to derive an approximate PDF for each data attribute and each simple hypothesis. These PDFs may be of any shape. Notably, they might be non-Gaussian. During the classification phase, a given datum is associated with a set of masses that are generated in an elaborated way from the aforementioned densities. Using the proposed contextual discounting method, mass functions are then weakened differently depending on the ability of each source to discriminate between classes. Mass functions of the different sources are then merged to have a consensual mass function using a suitable fusion rule. A final decision is then deduced using the so-called "pignistic transform" \cite{smets1994transferable}.

The rest of this paper is organized as follows: Section \ref{sec:rb} recalls the theoretical tools used in the proposed approach. Section \ref{sec:nsl} describes the NSL-KDD dataset. A description of Boosted PR-DS architecture is introduced in Section \ref{sec:bprds}. Section \ref{sec:er} discusses the experimental results by comparing them with those of some previous studies using the NSL-KDD dataset.  Final remarks and further suggestions for improvement are given in Section \ref{sec:cfw}.
	
	\section{Related background}
\label{sec:rb}
We succinctly outline some fundamentals of Dempster-Shafer theory, Parzen-Rosenblatt density estimation and contextual discounting. 

\subsection{Dempster-Shafer theory}
Suppose that $\Omega=\{\omega_1,...,\omega_K\}$, and  $\mathcal{P}(\Omega)=\{A_1,...,A_Q\}$ is its power set, where $Q=2^K$. A defined mass function $M$  ranging from $\mathcal{P}(\Omega)$ to $[0,1]$ is named a "basic belief assignment" (\textit{bba}) if $M(\emptyset)=0$ and $\sum_{A \in \mathcal{P}(\Omega)}{M(A)}=1$. A \textit{bba} $M$ therefore defines a ``plausibility'' function $Pl$ ranging from $\mathcal{P}(\Omega)$ to $[0,1]$ by $Pl(A)=\sum_{A\cap B\neq\emptyset}{M(B)}$, and a ``credibility'' function $Cr$ ranging from $\mathcal{P}(\Omega)$ to $[0,1]$ by $Cr(A)=\sum_{B \subset A}{M(B)}$. In addition, the two functions mentioned above are bound by $Pl(A)+Cr(A^c)=1$. Moreover, a probability function $p$ could be regarded as a particular case wherein $Pl=Cr=p$.

In case where two \textit{bba}s $M_1$ and $M_2$ denote two elements of evidence, we can combine them together using the ``Dempster-Shafer fusion'' (DS fusion), which results in $M=M_1\oplus M_2$ that is defined by:
\begin{equation}\label{eq:DSF}
M(A)=(M_1 \oplus M_2)(A)\propto \sum_{B_1\cap B_2=A}{M_1(B_1)M_2(B_2)}
\end{equation}

Lastly, through Smets’ technique\cite{smets1994transferable}, an evidential \textit{bba} $M$ can be converted into a probabilistic one, whereby every belief mass $M(A)$ is evenly distributed over all elements of $A$, resulting in the so-called ``pignistic probability'', $Bet$ , given by:

\begin{equation}\label{eq:pignistic}
Bet(\omega_i)=\sum_{\omega_i\in A\subseteq \Omega}{\frac{M(A)}{|A|}}
\end{equation}
where $|A|$ is the number of elements of $\Omega$ in $A$.

It is worth mentioning that there are various evidential fusion rules in the literature that deal differently with the issue of conflicting sources \cite{dubois1988representation}\cite{smets1990combination}\cite{sebbak2014evidential}.

\subsection{Parzen-Rosenblatt density estimation}
\label{sec:parz}
As a statistical tool, the Parzen-Rosenblatt window technique \cite{parzen1962estimation}\cite{rosenblatt1956remarks}, otherwise known as kernel density estimation, is a way to smooth data by making population inferences based on a finite sample. This technique can be perceived as a non-parametric method to construct the PDF $f$, of an unknown form, linked to a random variable $X$. Suppose $(x_1,x_2,...,x_N)$ an example of the realizations of such a random variable. The challenge is to estimate the $f$ values at multiple points of interest. The smoothing of the kernel can then be seen as a generalization of the histogram smoothing where a window, of a predetermined shape, centered at every point is utilized to approximate the value of density at the given point. This is done by using the following estimator:
\begin{equation}\hat{f}_h(x)=\frac{1}{Nh}\sum_{i=1}^{N}{K \left(\frac{x-x_i}{h} \right)}\end{equation} 
where $K(\cdot)$ is the kernel - a zero-mean non-negative function that integrates to one - and $h>0$ is a smoothing parameter known as ``kernel width''. Furthermore, it is possible to use a variety of kernel functions like Uniform (Box), Gaussian (Normal), Triangle, Epanechnikov \cite{epanechnikov1969non}, Quartic (Biweight), Tricube \cite{altman1992introduction}, Triweight, Logistic, Quadratic \cite{cleveland1988locally}, and others.

\subsection{Discounting methods}
Such methods can be used to estimate the weakening coefficients assigned to a source in order to correct its decision. These adjustments differ depending on whether it is a classic or contextual weakening.

\subsubsection{Classical discounting} 
The weakening of mass functions makes it possible to model sources' reliability by introducing a coefficient $\alpha^{s}$ where for each source $s$ we have:
\begin{equation}
\begin{cases}
m'^{s}(A)=\alpha^{s}.m^{s}(A)    &  \forall A\in 2^{\Omega }, A\neq \Omega      \\
m'^{s}(\Omega )=     (1- \alpha^{s}) +  \alpha^{s}. m(\Omega )
\end{cases}
\label{affaiblissement}
\end{equation}
$\alpha^{s}$ is the weakening coefficient of the  $s^{th}$ source.
Among the classical weakening methods, we find \cite{elouedi2004assessing} and  \cite{guo2006evaluating}.

\subsubsection{Contextual discounting}
The idea behind the contextual weakening is that the reliability of a source can vary depending on the truth of the object to be recognized (the context). For example, a sensor responsible for recognizing flying targets may be more or less able to discern certain types of aircraft. The method we propose belongs to this category and is described below.

\paragraph*{Weakening using F-score}
\label{para:weak}
In this method, we evaluate the ability of each attribute (source) to classify elements belonging to different hypotheses -simple or composite-. This is done by considering each attribute separately to classify a new element. Using a cross-validation process, a confusion matrix is obtained.
From this matrix, the "F-score" performance is calculated for all the hypotheses. These measures will be used as weakening coefficients and the equation \ref{CoefAffaib1} is applied to weaken the mass function of each source $s$.
\begin{equation}
\begin{cases}
m'^{s}\left( A \right)=\alpha_{A}^{s} m^{s}\left( A \right) \qquad  \qquad \hbox{$A\in \left\{ 2^{\Omega}/\Omega \right\}$}\\[3.2ex]
m'^{s}\left( \Omega \right)=m^{s}\left( \Omega \right) + \sum\limits_{A\in \left\{ 2^{\Omega}/\Omega \right\}}\left( 1-\alpha_{A}^{s} \right)m^{s}\left( A \right)
\end{cases}
\label{CoefAffaib1}
\end{equation}

$\alpha_{A}^{s}$ is the weakening coefficient of hypothesis A for the $s^{th}$ source.

\section{NSL-KDD dataset description}
\label{sec:nsl}
In addition to the fact that attack patterns are constantly evolving and changing, the challenge in building a robust Network Intrusion Detection System (NIDS) is that a real-time pattern of network data consisting of both intrusions and normal traffic is out of reach. This is why many recent works are still using the NSL-KDD dataset to evaluate the performance of their approaches \cite{gurung2019deep}\cite{yin2017deep}.

One of the most frequently used datasets for intrusion detection tests is the NSL-KDD dataset which was released in 2009 \cite{Tavallaee2009ADA}.
In addition to addressing efficiently redundant records' issue in the KDDCUP'99 dataset, NSL-KDD is designed by reducing the number of records in the training and test sets in a sophisticated manner to prevent the classifier from biasing towards frequent records. 

There are three datasets within NSL-KDD. One for training which is  KDDTrain+ and two for testing with an increasing difficulty respectively KDDTest+ and KDDTest-21, all of which having normal records as well as four distinct types of attack records, as shown in Table \ref{tab:classdistro}. KDDTest-21 which is a subset of the KDDTest+ is designed to be a more challenging dataset by removing the often correctly classified records. For more details about how KDDTest-21 was conceived, the reader may refer to \cite{Tavallaee2009ADA}.

\begin{table}[!h]
	\centering
	\caption{DIFFERENT CLASSES OF THE NSL-KDD DATASET.\label{tab:classdistro}}
	\begin{tabular}{cccccc}
		\hline
		& \textbf{Normal} & \textbf{Dos}   & \textbf{Probe} & \textbf{R2L}  & \textbf{U2R} \\ \hline
		\textbf{KDDTrain+}  & 67343  & 45927 & 11656 & 995  & 52  \\ \hline
		\textbf{KDDTest+ }  & 9711   & 7458  & 2421  & 2754 & 200 \\ \hline
		\textbf{KDDTest-21} & 2152   & 4342  & 2402  & 2754 & 200 \\ \hline
	\end{tabular}
\end{table}

Each record has 41 attributes and a class label as well. These attributes are divided into basic features, content features, and traffic features. Attacks in the dataset are grouped into four categories based on their characteristics: DoS (denial of service attacks), Probe (Probing attacks), R2L (root-to-local attacks) and U2R (user-to-root attacks). Some specific types of attacks are included in the test set but are not included in the training set. This makes it possible to provide a more realistic testing ground.

\section{Boosted PR-DS}
\label{sec:bprds}

This section describes the theoretical basis of the proposed intrusion detection scheme called Boosted Parzen-Rosenblatt Dempster-Shafer (Boosted PR-DS). To do this, suppose we have a sample of $N$ pre-tagged multiattribute data $(Z_1,...,Z_N)$ where each datum $Z_n=(X_n,Y_n)$ with $X_n\in \Omega=\{ \omega_1,...,\omega_K\}$ being the tag, and $Y_n=(Y_n^1,...,Y_n^P)\in \mathbb{R}^P$ being the $P$-attribute observation. The challenge is then to determine the tag of any new observation $Y_{n^\prime}$.

\begin{figure}[tbh!]
	\centering
	\includegraphics[width=0.8\linewidth]{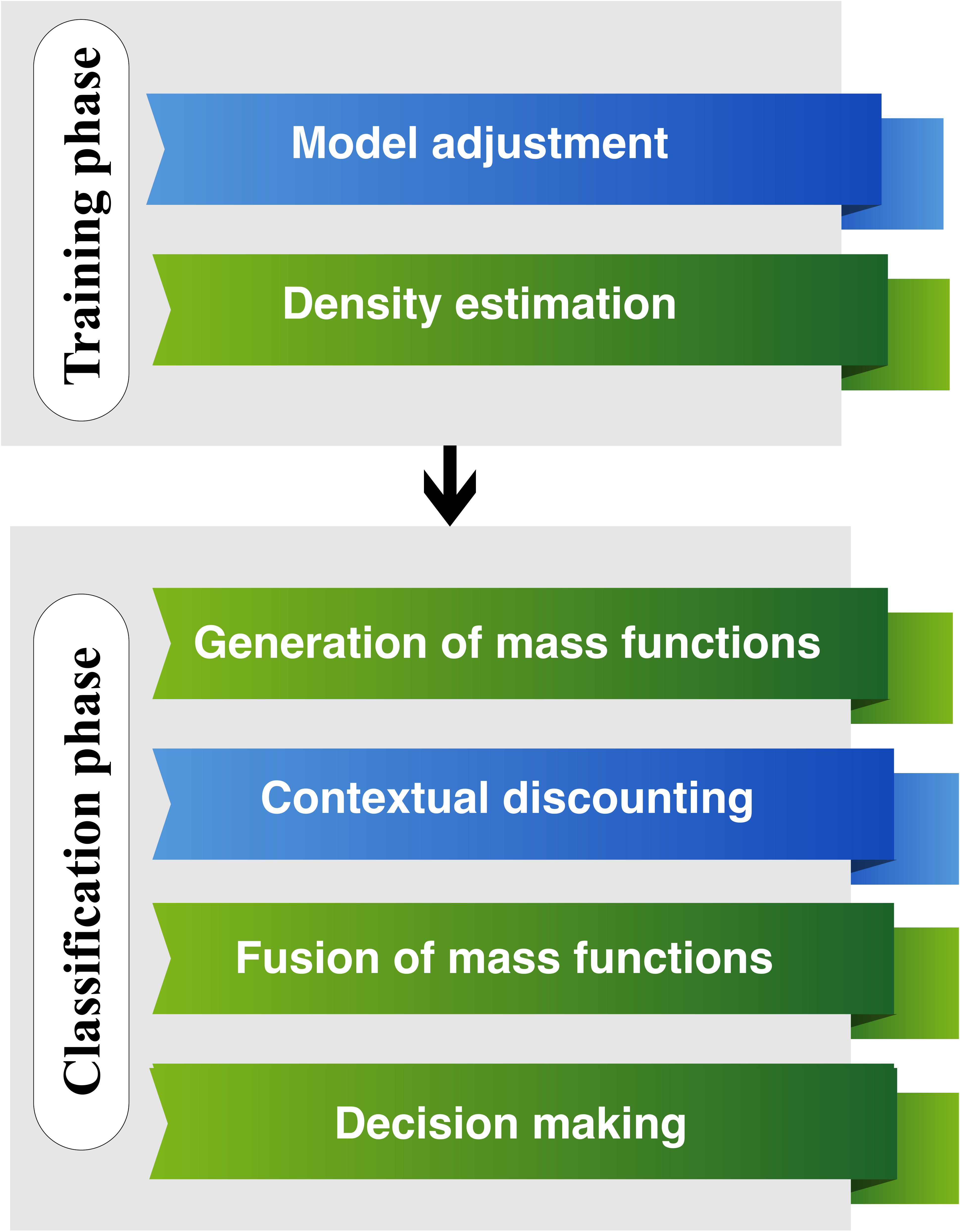}
	\caption{Proposed Boosted PR-DS framework}
	\label{fig:framework}
\end{figure}

As shown in Figure \ref{fig:framework}, we begin by briefly outlining the training process carried out on the pre-tagged data sample   $(Z_1,...,Z_N)$. Next, we illustrate the way our approach assigns a new observation $Y_{n^\prime}$ to one of the $K$ classes (tags).

%%%%%%%%%%%%%%%%%%%% This figure should be in Experimental results. Due to technical issues, I am putting it here %%%%%%%%%%%%%%%%%%%%%%%%%%%%%%
\begin{figure*}[tbh!]
	\centering
	\includegraphics[width=1\linewidth]{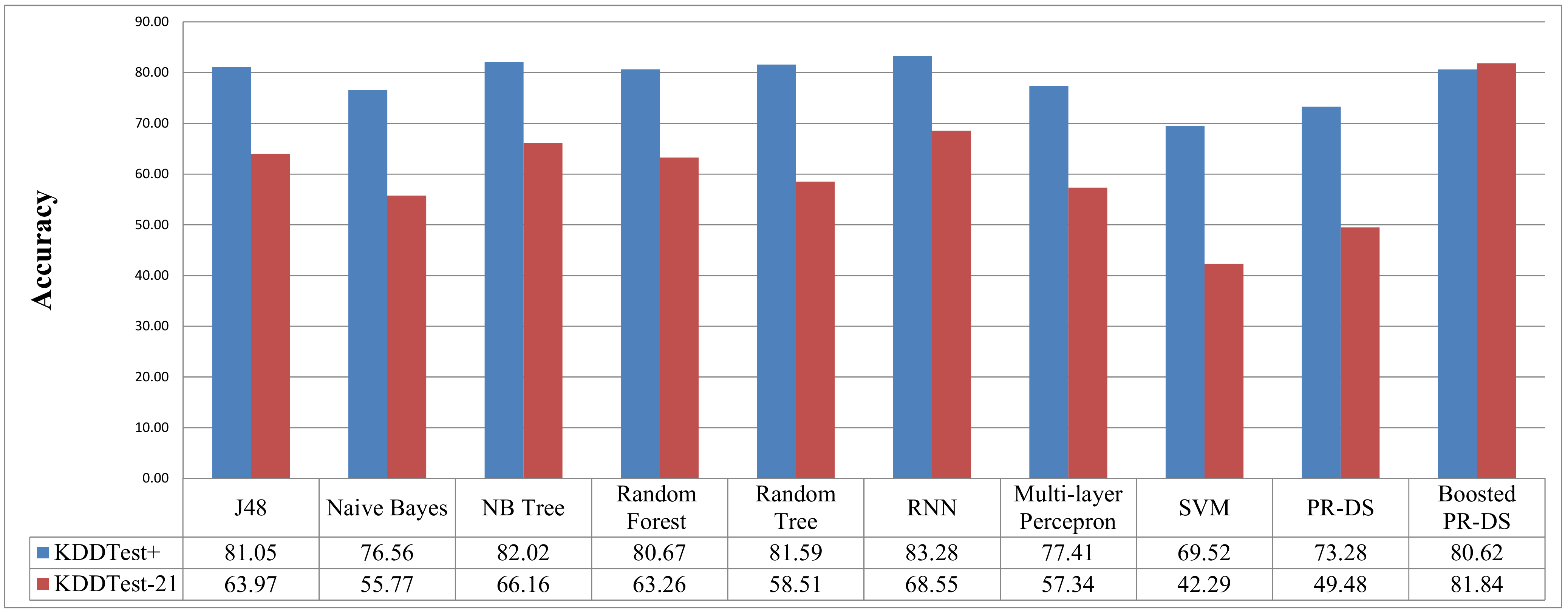}
	\caption{Performance of Boosted PR-DS and the other models on KDDTest+ and KDDTest-21.}
	\label{fig:preliminary-results}
\end{figure*}
%%%%%%%%%%%%%%%%%%%%%%%  end  %%%%%%%%%%%%%%%%%%%%%%%

\subsection{Training phase}

Consider the pre-tagged multi-attribute data above $(Z_1,...,Z_N)$. Under our Boosted PR-DS scheme, the training phase involves two steps. The first is model adjustment which consists of determining the optimal kernel and fusion rule for the data along with the computing of weakening coefficients for each hypothesis. The second step is density estimation where the previously chosen kernel is used to estimate the Probability Density Functions (PDF) of each class for all attributes. 
\subsubsection{Model adjustment}
In the first step, while changing kernels and fusion rules, basic PR-DS is used in a cross-validation process on the training data. The kernel and fusion rule giving the highest accuracy are then selected. To compute the weakening coefficients, we propose to use the F-score measures obtained from classifying each attribute (taken alone) as explained in paragraph Section  \ref{para:weak}.

\subsubsection{Density estimation}
In this step, we use the kernel chosen during the previous step to estimate densities using the Parzen-Rossenblatt method as described in Section \ref{sec:parz} instead of considering that they follow a normal distribution as in the classical case. We thus obtain, for each class $\omega_k\in \Omega$ and for each attribute $p$ ($1<p<P$), a Parzen-Rosenblatt density $\hat{f}^p_k$.

Eventually, in addition to the estimated densities, the trained model includes the weakening coefficients and the best-fit fusion rule.
\subsection{Classification phase}
Given a new observation $Y_{n^\prime}$, a mass function $M^p$ for each attribute is constructed based on the estimated densities. The proposed contextual discounting mechanism is then applied using the previously calculated weakening coefficients. Subsequently, the weakened mass functions are combined to obtain a consensual report $M$. The final decision is made using the so-called Pignistic Transform. In what follows, we describe these different steps.
\subsubsection{Generation of mass function}
In order to determine the mass $\mathcal{M}^p$ assigned to the attribute $p$, we will consider the rank function $\delta_p$ which is defined from $\{1,..,K\}$ to $\Omega$ so that $\delta_p(k)$ is the $k-$ranked element of $\Omega$ in terms of $\hat{f}^p$, i.e. $\hat{f}^p_{\delta_p(1)}(Y^p_{n^\prime})\leq \hat{f}^p_{\delta_p(2)}(Y^p_{n^\prime})\leq ... \leq \hat{f}^p_{\delta_p(K)}(Y^p_{n^\prime})$. Then, $\mathcal{M}^p$ is determined as follows:

\begin{equation} \label{eq:step1}
\left\{ \begin{array}{l} {\mathcal{M}^p(\Omega) \propto \hat{f}^p_{\delta_p(1)}(Y^p_{n^\prime}) } \\ \\ {\mathcal{M}^p( \{ \omega_{\delta_p(k)}, ..., \omega_{\delta_p(K)} \}) \propto   \hat{f}^p_{\delta_p(k)}(Y^p_{n^\prime})-\hat{f}^p_{\delta_p(k-1)}(Y^p_{n^\prime})} \end{array}\right.
\end{equation}

\subsubsection{Contextual discounting}
To fine-tune the ultimate mass assigned to the $p$ attribute, a weakening process based on the proposed contextual discounting mechanism mentioned in paragraph \ref{para:weak} is applied.

\subsubsection{Fusion of mass functions}
Mass Functions assigned to different attributes are then merged into a single consensus mass $M=\bigoplus_{p=1}^P{M^p}$ using the fusion rule selected on the training phase. 
\subsubsection{Decision making}
The final decision is made based on the Pignistic transformation of $M$:
\begin{equation}\label{eq:decision}
\hat{X}_{n^\prime}=\arg{\max_{\omega_k}{\sum_{A \ni \omega_k}{\frac{M(A)}{|A|}}}}
\end{equation}

It is worth noting that the novelty of Boosted PR-DS with respect to those using similar architectures is based on the steps of model adjustment, generation of mass function, and contextual discounting.

\section{Experimental results}
\label{sec:er}

To assess the performance of the proposed boosted PR-DS method, experimental tests are conducted on the NSL-KDD dataset containing two test sets of increasing difficulty, KDDTest+ and KDDTest-21 respectively, as described in Section \ref{sec:nsl}. 

A comparative analysis is made with regard to nine methods: J48 decision tree learning \cite{quinlan2014c4}, Naive Bayes \cite{john1995estimating}, NBTree\cite{kohavi1996scaling}, Random Forest \cite{breiman2001random}, Random Tree \cite{aldous1991continuum}, Multi-layer Perceptron \cite{ruck1990multilayer}, Support Vector Machine (SVM) \cite{chang2011libsvm}, and Recurrent Neural Networks (RNN) \cite{yin2017deep}, Parzen-Rosenblatt Dempster-Shafer (PR-DS) \cite{hamache2018uncertainty}. 

While giving a comparable accuracy on the KDDTest+ dataset, Boosted PR-DS outperforms the other state-of-the-art methods on the KDDTest-21 testing set as shown in Figure \ref{fig:preliminary-results}. This is mainly due to taking the estimated reliability into account by using the contextual discounting mechanism along with adjusting the model by selecting the most suitable kernel and fusion rule for a given training dataset.

\begin{figure}[tbh!]
	\centering
	\includegraphics[width=0.75\linewidth]{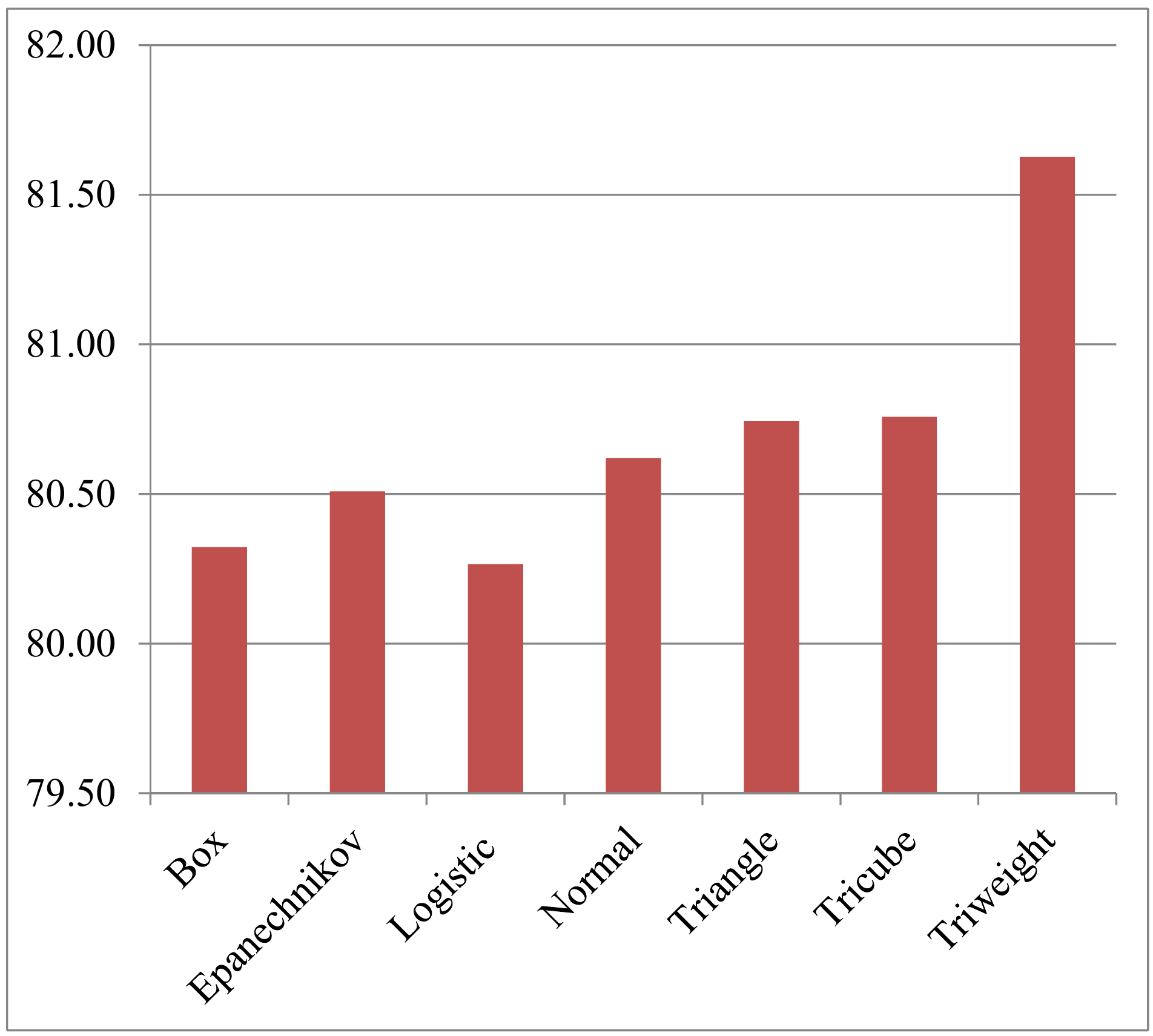}
	\caption{Boosted PR-DS on the KDDTest-21 dataset using different kernels}
	\label{fig:kernels-effect}
\end{figure}

To demonstrate the effect of kernel selection, we assess our approach on the KDDTest-21 dataset by changing the kernel each time, while maintaining the other parameters. Figure \ref{fig:kernels-effect} shows that three kernels at least are getting better results than the Normal kernel which confirms the relevance of choosing an adapted kernel to suitably constructing our densities instead of using the normality assumption.
%, height=0.25\textheight

\section{Conclusion and future work}
\label{sec:cfw}

As a conclusion, we can consider Boosted PR-DS as a combination of multiple classifiers where each attribute (source) is a classifier. By using contextual discounting, one may prioritize the decision of an individual classifier regarding those classes in which its accuracy was high in the training phase and be doubtful regarding those classes it did not classify well. Furthermore, Boosted PR-DS choose a suitable fusion rule to take advantage of each individual classifier's knowledge to achieve a consensus decision. Experimental results validate the interest of this approach with respect to other state-of-the-art intrusion detection models. As a possible future direction, it would be interesting to consider handling conflicting sources with a more sophisticated fusion rule.



	
	
	%\bibliographystyle{IEEEtran}
	% As suggested below, edit bibtemplate_samples.bib to reflect
	% your bibliography. See bibtemplate.text for referencing.
	%
	
	\bibliographystyle{IEEEtran}
	\bibliography{Mybib}

\end{document}